\documentclass[conference]{IEEEtran}

\usepackage{tikz}
\usepackage{cite}
\usepackage{amsmath,amssymb,amsfonts}
\usepackage{comment}
\usepackage{algorithmic}
\usepackage{graphicx}
\usepackage{url}
\usepackage{textcomp}
\usepackage{xcolor}
\usepackage{colortbl}
\PassOptionsToPackage{hyphens}{url}
\usepackage{hyperref}
\usepackage{authblk}
\usepackage[inline]{enumitem}
\def\BibTeX{{\rm B\kern-.05em{\sc i\kern-.025em b}\kern-.08em
    T\kern-.1667em\lower.7ex\hbox{E}\kern-.125emX}}
\begin{document}

\title{\fontsize{22pt}{18pt}\selectfont EmoTale: An Enacted Speech-emotion Dataset in Danish}



\author[2,3\IEEEauthorrefmark{1}]{Maja J. Hjuler}
\author[1]{Harald V. Skat-Rørdam}
\author[1]{Line H. Clemmensen}
\author[1\textdagger]{Sneha Das}
\affil[1]{Dept. of Applied Mathematics and Computer Science, Technical University of Denmark, 2800 Lyngby, Denmark}
\affil[2]{University Grenoble Alpes, CNRS, Grenoble INP, LIG, 38000 Grenoble, France } 
\affil[3]{School of Computer Science, Queensland University of Technology, Brisbane QLD 4000, Australia \authorcr Email: {\tt maja-jonck.hjuler@univ-grenoble-alpes.fr, \{harsk, lkhc, sned\}@dtu.dk} \vspace{-1ex}}

\maketitle
\begingroup\renewcommand\thefootnote{\IEEEauthorrefmark{1}}
\footnotetext{The author was affiliated with the Technical University of Denmark when this work was carried out.}
\endgroup
\begingroup\renewcommand\thefootnote{\textdagger}
\footnotetext{Corresponding Author}
\endgroup

\begin{abstract}
While multiple emotional speech corpora exist for commonly spoken languages, there is a lack of functional datasets for smaller (spoken) languages, such as Danish. 
To our knowledge, {\it Danish Emotional Speech~(DES)}, published in 1997, is the only other database of Danish emotional speech. We present EmoTale\footnote{Link to the dataset and source code: \url{{https://github.com/snehadas/EmoTale}}}; a corpus comprising Danish and English speech recordings with their associated enacted emotion annotations.
We demonstrate the validity of the dataset by investigating and presenting its predictive power using speech emotion recognition (SER) models. We develop SER  models for EmoTale {\it and} the reference datasets using self-supervised speech model (SSLM) embeddings and the openSMILE feature extractor. We find the embeddings superior to the hand-crafted features. The best model achieves an unweighted average recall (UAR) of 64.1\% on the EmoTale corpus using leave-one-speaker-out cross-validation, comparable to the performance on DES. 
\end{abstract}


\section{Introduction \& Background}
Speech signals are rich in information, both linguistic (in the form of sentences and words) and paralinguistic (denoting mood and affective state). Speech also carries information about multiple, potentially personal traits of the speaker, such as age, gender, and nationality.
Multiple psychological and neuroscientific models of the mind hypothesize that language and emotion are certainly linked~\cite{lindquist2017a}. For example, some cultures express anger more vocally, while others might be more restrained. Investigating voice and speech to judge emotional states dates back more than half a century~\cite{fairbanks1938vocal, soskin1961judgment}, and the earliest speech emotion recognizers (SERs) were proposed over two decades ago~\cite {dellaert1996recognizing, schuller2018speech}. 

\begin{figure}[!tbh]
\centering
\includegraphics[width=.95\columnwidth]{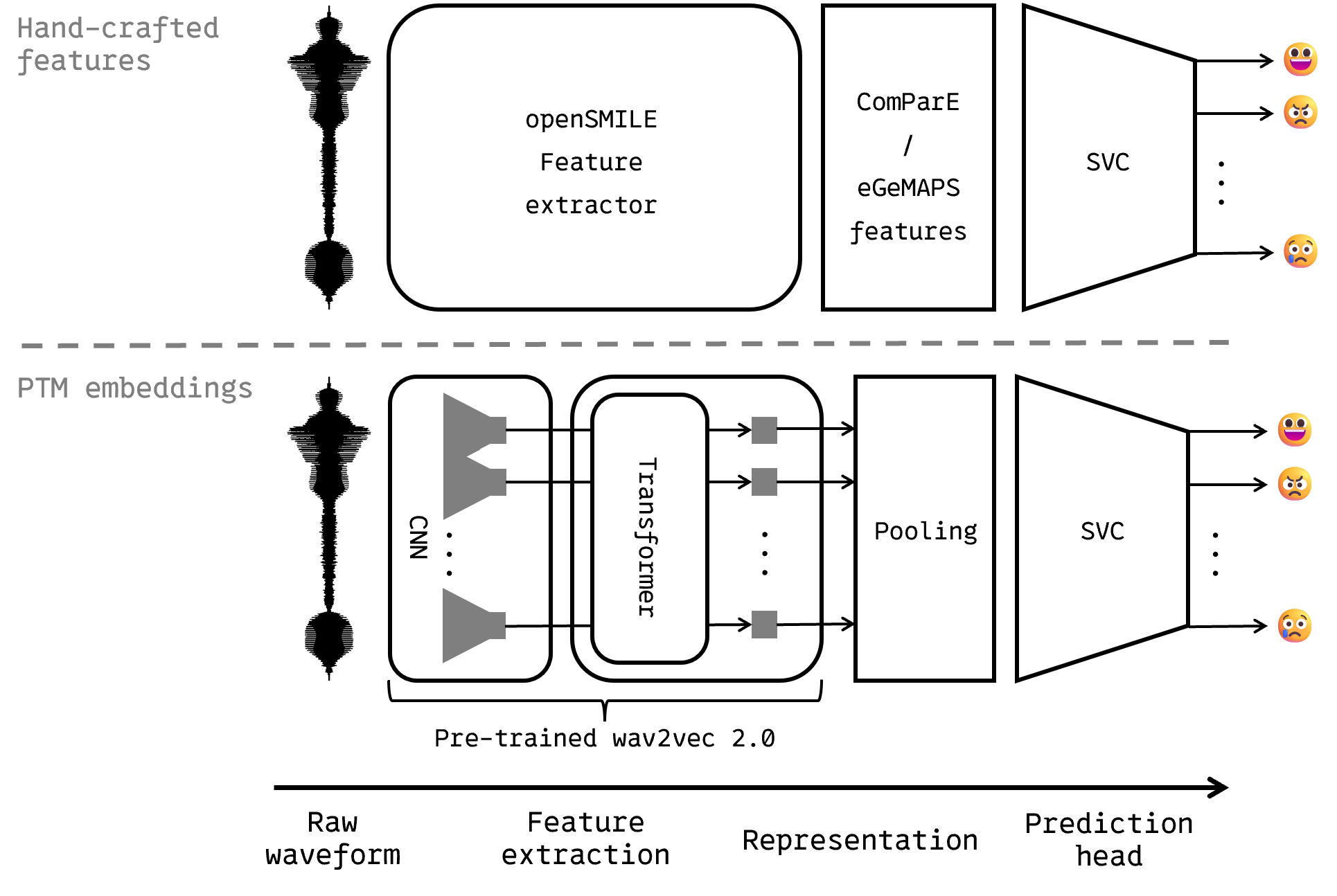}
\vspace{-.3cm}
\caption{Processing pipelines for hand-crafted (top) and deep features (bottom).} 
\label{fig:pipeline}
\vspace{-0.5cm}
\end{figure}
Emotions are inherently subjective; different people perceive emotions differently, and this can lead to differences in annotating emotional data~\cite{kang2023praxes, yannakakis2017ordinal}. 
Overall, two different labeling schemes are adopted in the literature: categorical class labels, which are nominal and discrete, and dimensional labels, which are continuous. 
The former often follows the {\it basic emotion theory} developed by Paul Ekman~\cite{Ekman_1992}, which assumes the existence of six basic and universal emotions that transcend language, cultural, and ethnic differences. 
The emotions, also known as {\it The Big 6}, are anger, disgust, fear, happiness, neutral, and sadness. 
Following the dimensional scheme, emotions can be described numerically in the two dimensions \textit{activation/arousal} and \textit{valence}, or in three dimensions by including \textit{dominance}. 
For example, happiness is characterized by positive valence, high activation, and neutral dominance, i.e., neither dominant nor submissive. 

In speech emotion recognition~(SER), frequently used emotions include happiness, anger, sadness, disgust, fear, frustration, surprise, and boredom. 
For a baseline comparison, it is common practice to include neutral as one of the emotions expressed. 
In many SER databases, utterances are spoken with \textit{enacted} emotions, but emotional responses can also be induced through specific tasks, scenarios, or stimuli to capture genuine emotional expressions. 
Alternatively, \textit{natural/spontaneous} speech can be collected from existing digital resources, such as TV shows or podcasts, and annotated retrospectively. 
For English SER, IEMOCAP~\cite{Busso_2008_IEMOCAP} and MSP-Podcast~\cite{Lotfian_2019, busso2016msp} are two of the most frequently used corpora due to their relatively large size, and the inclusion of both categorical and dimensional labels.
The Danish DES database~\cite{Engberg_1997} was published in 1997 and contains four speakers (two male and two female) expressing five emotions: neutral, surprise, happiness, sadness, and anger.
All utterances are equally balanced for each gender and actor.
In listening tests for the DES corpus, emotions were correctly classified 67.3\% of the time on average~\cite{Engberg_1997}.
However, DES includes single words and questions, and it was not developed specifically for speech emotion recognition purposes.

Contemporary state-of-the-art SER research is most often based on deep learning models~\cite{Wagner_2023, Wang_2022, Pepino_2021, zhang_2021} like the SUPERB benchmark~\cite{Yang_2021_SUPERB}. Rapid development of scale-based deep learning was enabled by the availability of large and exhaustive speech emotion datasets~\cite{Busso_2008_IEMOCAP, Lotfian_2019}. 
The most comprehensive SER datasets are in English or other large~(spoken) languages. Developing SER models that transfer well to unseen languages, addresses the lack of resources in smaller languages while enabling the accessibility of these models. However, at minimum, a test dataset is necessary to validate the suitability and safety of a SER model before deployment. In this work, we take the first step towards presenting a Danish-SER dataset to address the gap in functional datasets. Our {\it contributions} are: \begin{enumerate*}
\item the EmoTale dataset: a corpus comprising 450 Danish and 350 English speech recordings with associated categorical and dimensional emotion annotations.
\item we also present a thorough validation of the quality of EmoTale by analyzing its predictive capacity using reference datasets. Through this process, we revisit {\it transferability of SER} and present insights with respect to other multilingual datasets. \end{enumerate*}

\section{Design of EmoTale}
To enable cross-corpus comparability and transferability, the design choices in EmoTale are similar to existing small-scale SER datasets.
The data collection procedure was inspired by the Berlin Database of Emotional Speech (Emo-DB)~\cite{Burkhardt_2005}. 

\subsection{Dataset curation}
\noindent
{\bf Recruitment: }Participants with acting experience and Danish {\it and} English language skills were recruited through physical flyers and posts on social media, and theater schools in the Greater Copenhagen area were contacted by email and phone. 
An online registration form was available in Google Forms, where participants signed up by providing their contact information and choosing their desired experiment date from a list of options. 
The exclusion criteria were age $<7$ years {\it or} no Danish-speaking skills. In compliant with GDPR requirements, we obtained written consent from the participant or the guardian of participants under $18$, and information about gender and age was recorded. 

{\renewcommand{\arraystretch}{1.2}
\begin{table}[!t]
\resizebox{.99\columnwidth}{!}{%
\begin{tabular}{|p{0.2cm}|p{4.5cm}|p{4.5cm}|}
\hline
\textbf{No.} & \multicolumn{1}{c|}{\textbf{Danish sentence}}                       & \multicolumn{1}{c|}{\textbf{English sentence}}                           \\ \hline
1.           & Dugen ligger på køleskabet.                                         & The tablecloth is lying on the fridge.                                   \\
2.           & Det sorte ark papir er placeret deroppe ved siden af tømmerstykket. & The black sheet of paper is located up there beside the piece of timber. \\
3.           & De bar det bare ovenpå og nu skal de ned igen.                      & They just carried it upstairs and now they are going down again.         \\
4.           & Det vil være på det sted, hvor vi altid opbevarer det.              & It will be in the place where we always store it.                        \\
5.           & Om syv timer er det morgen.                                         & In seven hours it will be morning.                                       \\ \hline
\multicolumn{3}{|c|}{\textbf{Five emotions: Neutral, Anger, Sadness, Happiness, Boredom}}\\
\hline
\end{tabular}
}
\vspace{0.2cm}
\caption{Danish and English sentences in EmoTale.}
\vspace{-0.8cm}
  \label{tab:sentences}
\end{table}}
\noindent
{\bf Data collection procedure:}
The data recordings were performed in multiple sessions and locations with no ambient noise.
At the start of a session, the participant was ﬁtted with RØDE Wireless Go microphones and was walked through the experiment, and allowed to ask questions. Five sentences were enacted with five different emotions~(Tab.~\ref{tab:sentences}), and the participant enacted all sentences for a specific emotion before moving on to the next. The sentences are translations of selected sentences from Emo-DB~\cite{Burkhardt_2005}; to minimize subjective associations and differences, the sentences were selected such that they are emotionally neutral and comprise minimal contextual information. We relied on the participant’s ability to self-induce an emotion by recalling a situation where it had been felt strongly. The participants were allowed to repeat the sentences as many times as they liked, but only the last recording was retained. 
Since Danish speakers are fluent in English, the participants could choose to contribute with enacted English speech in addition to Danish.
The utterances were recorded at a 48 kHz sampling frequency and saved in .WAV format. 
The audio filenames comprise the meta information on the language, speaker ID, emotion, and sentence. For example, the file \texttt{DK\_004\_A\_5.wav} is the fifth sentence spoken by speaker \texttt{004} in Danish, with {\it angry} affect.\\
\noindent
{\bf Data protection and ethical considerations:} Ethical approval was obtained from the institutional review board prior to the study~\cite{IRB_emotale}. The samples are pseudo-anonymized by generating a random identifier for each participant. Since the emotions are enacted and the selected sentences do not contain personal contextual information, potential misuse of the data to cause harm to the participants is reduced. The dataset is supported by a datasheet~\cite{gebru2021datasheets}, in the later part of the paper.\\
{\bf Annotation procedure: }
In addition to emotion categories, many existing datasets annotate speech-emotion samples using dimensional labels~\cite{Busso_2008_IEMOCAP, das_2022_III}. We adopt a similar approach in the EmoTale corpus, where utterances are manually annotated for arousal, valence, and dominance on a scale from 1 to 5, with increments of 0.5. Arousal indicates the level of excitement or activation associated with the emotion, ranging from calm (1) to excited (5). Valence reflects the emotional tone, with values ranging from negative (1) to positive (5). Dominance measures the level of dominance associated with the emotion, with a scale from submissive (1) to dominant (5). The first, second, and last authors independently assigned labels to all utterances in EmoTale, each providing one categorical label for the intended emotion and three numerical labels for arousal, valence, and dominance. The categorical annotations serve to validate the enacted emotions. 

\begin{figure}[!t]
\centering
\vspace{-.3cm}
\includegraphics[width=0.9\columnwidth]{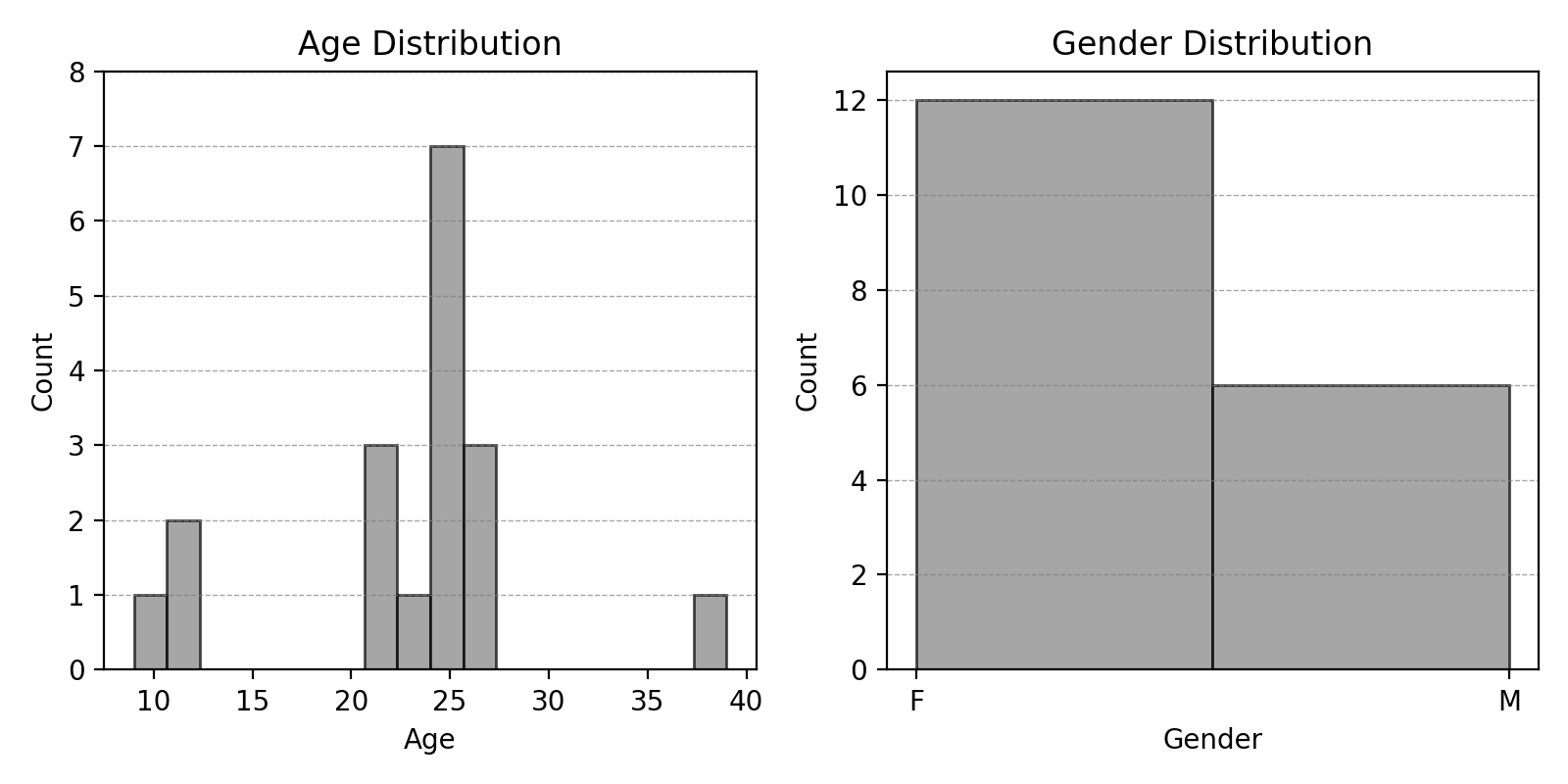}
\vspace{-.3cm}
\caption{Age and gender distribution of EmoTale participants.}
\label{fig:age_gender_dist}
\vspace{-.4cm}
\end{figure}
\noindent 

\subsection{Description of EmoTale}
\noindent
EmoTale comprises emotional speech from 18 participants, of whom 12 are female and six are male. 
The total number of Danish and English utterances are 450 and 350, respectively. The average age of the participants was 22.8 years, ranging from 9 to 39 years old. Age and gender distributions of participants are illustrated in Fig.~\ref{fig:age_gender_dist}. The goal of this dataset is to develop infrastructure to enable the evaluation and safe deployment~\cite{das2022speech} of existing speech processing and SER on the Danish-speaking population, including children. Therefore, speakers under the age of 18 are also included in the dataset. Some files were cropped to exclude a ‘click’ sound~(from experimenters' keyboard) at the start or end of the recording. 

\noindent
{\bf Inter-rater reliability (IRR):}
In addition to the enacted emotion, three independent annotators provided four labels per instance: one categorical label for the intended emotion and three numerical labels for arousal, valence, and dominance, each ranging from 1 to 5 with increments of 0.5. Valence [1-negative, 5-positive], activation [1-calm, 5-excited], and dominance [1-weak, 5-strong]. 
We employ Cohen's Kappa ($\kappa$)~\cite{Cohens_kappa} to assess inter-rater reliability (IRR) between the categorical labels provided by the first two annotators, as well as to evaluate their agreement with the predefined ground truth emotion. The IRR results are presented in Table~\ref{tab:Cohens_kappa_IRR}. A value of $\kappa=1$ implies perfect agreement, and $\kappa=0$ means the agreement is exactly what would be expected by chance. $0.7<\kappa$ indicates good to substantial agreement~\cite{IRR}.
To evaluate the IRR between dimensional emotion annotations (valence, arousal, dominance), we employ Concordance Correlation Coefficient (CCC)~\cite{lawrence1989concordance}, which is suitable for ratings on a fine-grained, continuous, or interval scale. As seen in Table~\ref{tab:CCC}, the results indicate moderate to strong agreement for arousal and valence, and moderate agreement for dominance. 

{\renewcommand{\arraystretch}{1.2}
\begin{table}[!t]
\centering
\begin{tabular}{c|c|c|c}
\hline
\textbf{} & \textbf{a1 vs.~a2} & \textbf{a1 vs.~GT} & \textbf{a2 vs.~GT} \\ \hline
$\kappa$  & 0.71                    & 0.75                    & 0.85                    \\ \hline
\end{tabular}
\label{tab:Cohens_kappa_IRR}
\vspace{0.1cm}
\caption{Cohen's Kappa reliability between categorical labels from annotators 1 and 2 (a1, a2) and the predefined emotion (GT).}
\vspace{-0.3cm}
\end{table}}

{\renewcommand{\arraystretch}{1.2}
\begin{table}[!t]
\centering
\begin{tabular}{c|c|c|c}
\hline
\textbf{} & \textbf{Arousal} & \textbf{Valence} & \textbf{Dominance} \\ \hline
CCC       & 0.72             & 0.75             & 0.57               \\ \hline
\end{tabular}
\vspace{0.1cm}
\caption{Concordance Correlation Coefficient (CCC) between dimensional labels from annotators 1 and 2.}
\label{tab:CCC}
\vspace{-0.8cm}
\end{table}}

\section{Validating the {\it emotion}-signal in EmoTale}
We validate the signals in EmoTale by a) comparing human annotations to the predictions from a pre-trained SSL, and b) analyzing the predictive power of the data samples by training and evaluating SER models in Danish. We employ the following datasets as references on the validity and quality of EmoTale: Emo-DB~(German), Urdu~(Urdu)~\cite{latif2018cross}, DES~(Danish), and AESDD~(Greek)~\cite{vryzas2018speech}.
\begin{figure}[!tbh]
\centering
\includegraphics[width=0.9\columnwidth]{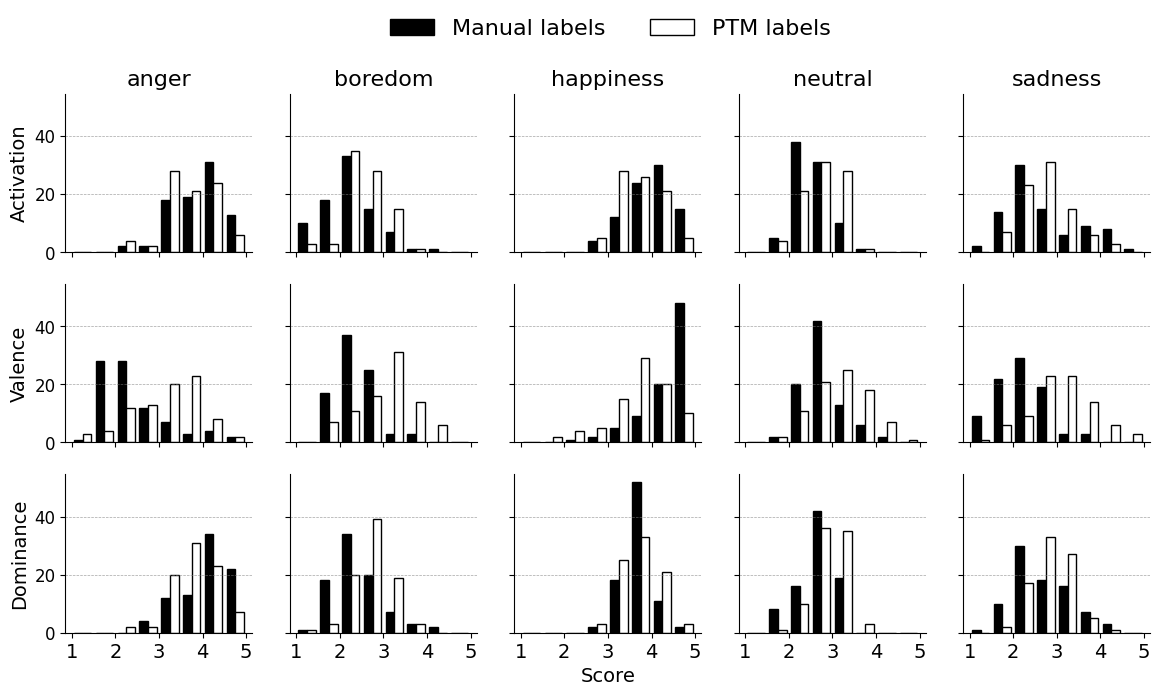}
\caption{EmoTale Annotator 1 labels for the utterances in Danish compared to the dimensional labels computed from PTM output.}
\label{fig:ptm_versus_label_distribution}
\end{figure}
\begin{figure}[!tbh]
\centering
\includegraphics[width=.9\columnwidth]{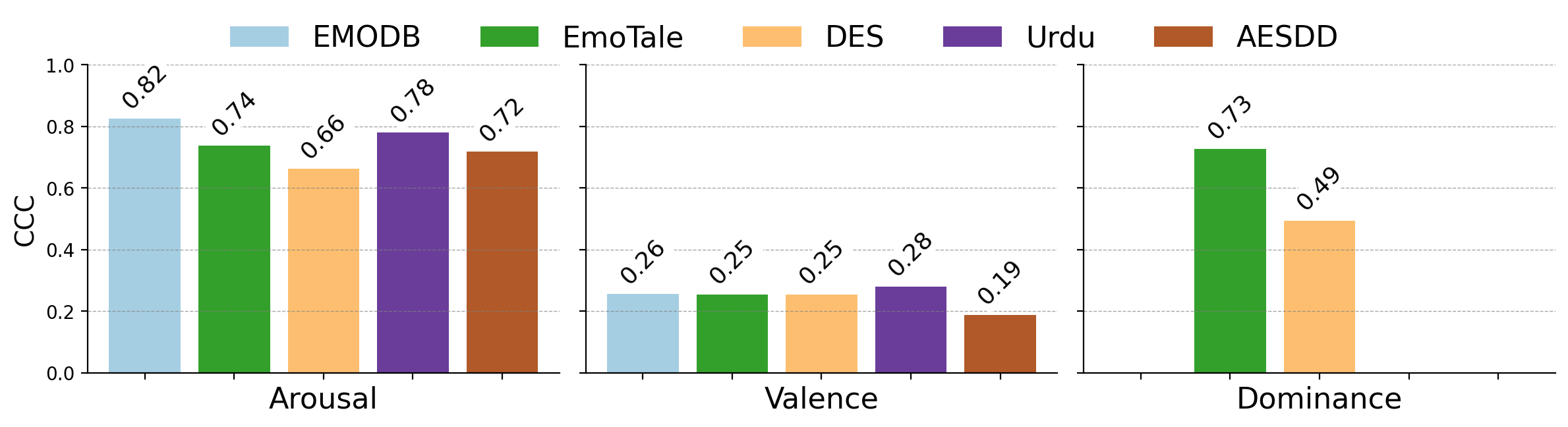}
\caption{Concordance Correlation Coeﬃcients between \texttt{w2v2-FT-dim} and manual labels across different datasets. Dominance labels are only available for DES and EmoTale.}
\label{fig:ccc}
\vspace{-0.5cm}
\end{figure}

\subsection{Labels: Human-annotation vs.~Pre-Trained Model}
A pre-trained model~(PTM), \texttt{\texttt{w2v2-FT-dim}}, fine-tuned on the MSP-Podcast with dimensional labels\footnote{\scriptsize\url{https://huggingface.co/audeering/wav2vec2-large-robust-12-ft-emotion-msp-dim}}, outputs activation, valence, and dominance scores ranging between 0 to 1. 
These were rescaled to a range between 1 and 5 to compare with manual labels as follows:
\begin{equation}
    A_{i, \text{scaled}} = 1 + 4 \frac{A_i - A_\text{min}}{A_\text{max} - A_\text{min}}, 
\end{equation}
where $A_i$ denotes the activation score, and $A_{\text{min}}$ and $A_{\text{max}}$ are the overall minimum and maximum activation scores, respectively. 
Valence and dominance scores were rescaled in the same way. 
Fig.~\ref{fig:ptm_versus_label_distribution} compares the scores predicted by the PTM for EmoTale to the labels by Annotator 1. 
Activation and valence labels for Emo-DB, Urdu, and AESDD were employed using~\cite{das2022continuous}, while DES and EmoTale were annotated as part of this work. Predictions by \texttt{w2v2-FT-dim} were compared against the human-annotated labels using CCC in Fig.~\ref{fig:ccc}; A high CCC is observed for activation/arousal and dominance, implying a high agreement between the outcome of PTM and human-annotated labels, but the scores are consistently lower for valence over all datasets.

\subsection{Validation with handcrafted features \& PTM embeddings}
We explore the predictive power of the samples in EmoTale with respect to the reference datasets by evaluating the performance of SER models on all the datasets, in the process revisiting cross-lingual transferability of speech-emotions. \\
\noindent
{\bf Method: }As for the SER models, we employ a support vector classifier~(SVC), a) with hand-crafted features, and b) PTM embeddings, also known as deep features. 
The PTM feature embeddings are extracted as the last hidden states of the pre-trained model, i.e.,~the last layer before any task-specific head is applied, and it is assumed that model embeddings provide a compact representation of the emotional content in a speech signal.
In transformer models, this is the output from the final transformer block. 
The experimental procedure is adapted from the one outlined by Wagner et al\footnote{\scriptsize\url{https://github.com/audeering/w2v2-how-to/blob/main/notebook.ipynb}}. 
The speech samples in DES, EmoTale, Emo-DB, Urdu, and AESDD datasets are downsampled to 16 kHz as the PTM input requirement, and stereo audio files were converted to mono by averaging to a single channel. The pipelines are illustrated in Fig.~\ref{fig:pipeline}. 
 
The eGeMAPS (extended Geneva Minimalistic Acoustic Parameter Set)~\cite{eyben2015geneva} and the ComParE (Computational Paralinguistics Challenge)~\cite{schuller2010interspeech} feature sets were extracted using the openSMILE toolkit~\cite{eyben2010opensmile} and serve as two separate baselines. 
These were tested against embeddings from the wav2vec2 base model\footnote{\scriptsize\url{https://huggingface.co/facebook/wav2vec2-base}}~\cite{Baevski_2020} as well as a wav2vec2 model fine-tuned for SER on the RAVDESS corpus~\cite{livingstone_2018_Ravdess} (\texttt{w2v2-FT-cat})\footnote{\scriptsize\url{https://huggingface.co/ehcalabres/wav2vec2-lg-xlsr-en-speech-emotion-recognition}} and one fine-tuned on MSP-Podcast~(\texttt{w2v2-FT-dim})~\cite{wagner_2022_wav2vec2}. 
The latter is fine-tuned on dimensional scores and not categorical labels, therefore, the output of hidden states is necessary to access the latent space of the model. 
Model embeddings are extracted by applying average pooling over the hidden states of the last transformer layer.
Subsequently, the features are input to a SVC with a linear kernel, and Leave-One-Speaker-Out (LOSO) cross-validation is applied. 
In each fold, features were standardized using the mean and standard deviation of the respective training set. We used a linear kernel to resemble the method in~\cite{schuller_2016_ComParE}. \\
\noindent
{\bf Evaluation: }
Applying LOSO cross-validation introduces variability in the performance metric. 
The aggregated unweighted average recall (UAR) across cross-validation folds is used for evaluation. 
However, it may overlook performance differences across individual speakers.  
Each iteration of LOSO involves training a model on a different subset of data, hence, for $S$ speakers it is more accurate to consider the $S$ different models separately.
For the same dataset, each model is tested under the same conditions, 
whereby we can apply paired t-tests to statistically model performances.
The UAR scores are computed as the sum of class-wise recall divided by the number of emotion classes, and the overall score is the average UAR across all datasets.
To provide a comprehensive view of model performance, we report both the aggregated results (highlighted rows in Tab.~\ref{tab:RQ1_results}) and the mean results across speakers (Speaker UAR). 
The former combines the predictions of all folds into a single confusion matrix and calculates the UAR. 
Once the SVC parameters are ﬁxed, changing the random seed does not affect results, hence, the standard deviation is zero.
The latter calculates the UAR for each LOSO cross-validation fold individually and takes the mean to consider how well the model generalizes across speakers. 
Similarly, sentence UARs are found by first grouping prediction sentences, calculating the UAR per group, and then taking a simple average across the groups.
In this way, all the speakers and sentences are given equal weight.
Standard deviations are reported to provide insights into the variability of the UAR scores across speakers, sentences, and datasets.
Sentence UARs are not included for the Urdu corpus since it contains natural utterances, hence, no sentences are repeated. 
\noindent
{\bf Results: }
For Emo-DB, the results reported in Table~\ref{tab:RQ1_results} using ComParE and \texttt{w2v2-FT-dim} embeddings are reproduced from~\cite{Wagner_2023}.
The performance trends observed on EmoTale align with those seen in Emo-DB and DES, reinforcing the consistency and reliability of the dataset. Specifically, the UAR scores for the three datasets follow the same trend with model performance in descending order using the features: \texttt{w2v2-FT-dim}, \texttt{w2v2-FT-cat}, ComParE, eGeMAPS, and \texttt{w2v2-b}.
Interestingly, Urdu deviates from the trend with eGeMAPS features outperforming both ComParE and the embeddings from the PTM fine-tuned on categorical labels, \texttt{w2v2-FT-cat}. 
In all cases, deep features from the ﬁne-tuned models yield the highest UARs, while embeddings from the wav2vec2 model without ﬁne-tuning perform the worst. 
Furthermore, the PTM fine-tuned on dimensional labels leads to the highest mean UAR across datasets, highlighting the benefit of ﬁne-tuning.

{\renewcommand{\arraystretch}{1.2}
\begin{table}[t]
\resizebox{\columnwidth}{!}{%
\begin{tabular}{lcccccc}
\hline
Corpus               & \textbf{Emo-DB} & \textbf{DES} & \textbf{EmoTale} & \textbf{Urdu} & \textbf{AESDD} & Overall     \\
\#Speakers           & 10              & 4            & 18               & 22            & 6              &             \\ \hline
\rowcolor[HTML]{C0C0C0} 
\textbf{ComParE}     & 79.0            & 48.5         & 52.0             & 50.0          & 58.0           & 57.5 ± 11.2 \\
Speaker UAR          & 74.9 ± 7.9      & 48.5 ± 10.5  & 50.9 ± 11.1      & 49.7 ± 41.5   & 58.2 ± 10.9    &             \\
Sentence UAR         & 79.5 ± 5.0      & 48.5 ± 10.1  & 52.0 ± 1.5       & -             & 58.1 ± 5.4     &             \\ \hline
\rowcolor[HTML]{C0C0C0} 
\textbf{eGeMAPS}     & 64.3            & 42.7         & 46.0             & 58.0          & 47.6           & 51.7 ± 8.1  \\
Speaker UAR          & 60.3 ± 14.1     & 42.7 ± 13.6  & 44.8 ± 13.8      & 29.2 ± 21.6   & 47.8 ± 11.8    &             \\
Sentence UAR         & 63.9 ± 5.5      & 42.7 ± 9.0   & 46.0 ± 3.9       & -             & 47.7 ± 7.1     &             \\ \hline
\rowcolor[HTML]{C0C0C0} 
\textbf{w2v2 base}   & 58.9            & 32.7         & 29.7             & 33.5          & 41.8           & 39.3 ± 10.6 \\
Speaker UAR          & 56.7 ± 8.0      & 32.7 ± 4.2   & 29.4 ± 6.9       & 23.1 ± 26.7   & 41.7 ± 12.3    &             \\
Sentence UAR         & 58.4 ± 8.2      & 32.7 ± 8.8   & 29.8 ± 2.6       & -             & 41.8 ± 8.5     &             \\ \hline
\rowcolor[HTML]{C0C0C0} 
\textbf{w2v2 FT dim} & \textbf{96.1†}  & 67.7         & \textbf{64.1‡}   & 59.5          & \textbf{83.2‡} & 74.1 ± 13.6 \\
Speaker UAR          & 94.7 ± 3.9      & 67.7 ± 4.0   & 62.0 ± 12.4      & 48.4 ± 36.5   & 83.1 ± 7.8     &             \\
Sentence UAR         & 96.1 ± 3.0      & 67.7 ± 12.7  & 64.1 ± 5.4       & -             & 83.2 ± 5.9     &             \\ \hline
\rowcolor[HTML]{C0C0C0} 
\textbf{w2v2 FT cat} & 88.8            & 62.7         & \textbf{59.6‡}   & 52.5          & \textbf{77.5‡} & 68.2 ± 13.1 \\
Speaker UAR          & 88.1 ± 5.0      & 62.7 ± 5.1   & 57.8 ± 12.2      & 37.1 ± 34.7   & 77.5 ± 10.8    &             \\
Sentence UAR         & 88.3 ± 4.8      & 62.7 ± 10.5  & 59.6 ± 3.5       & -             & 77.6 ± 8.1     &             \\ \hline
\end{tabular}
}
\vspace{0.1cm}
\caption{UAR (\%) for SVC based on hand-crafted \& deep features as mean and std.~dev.~($\mu \pm \sigma$) over LOSO folds. 
The PTMs are a base model (\texttt{w2v2-b}) and models ﬁne-tuned on dimensional labels (\texttt{w2v2-FT-dim}) and categorical labels (\texttt{w2v2-FT-cat}). UAR scores across speakers and sentences provide insights into the performance variability, except for Urdu, which contains natural speech. † and ‡ mark the single best and the two
best models across LOSO folds with statistical signiﬁcance for a dataset.}
\label{tab:RQ1_results}
\vspace{-0.8cm}
\end{table}}


To further validate model performance on EmoTale, we applied \textit{pairwise} t-tests~\cite{brockhoff2018introduction} across LOSO folds to assess the statistical significance of differences between feature sets.
While fine-tuning of dimensional labels (\texttt{w2v2-FT-dim}) yields a statistically significant improvement over categorical labels (\texttt{w2v2-FT-cat}) for Emo-DB, this distinction does not hold for EmoTale nor the other datasets, which negates the argument against categorical labels~\cite{das2022towards}. 
Similarly, for several datasets, there is no statistically significant difference in model performance when training on eGeMAPS features compared to wav2vec2 base model embeddings. 
The single best and two best models with statistical significance are marked in Table~\ref{tab:RQ1_results} when such a conclusion could be drawn based on pairwise t-tests.
These findings further strengthen EmoTale’s role as a reliable benchmark for emotional speech, with results that reflect those of established corpora.


We also observe from Tab.~\ref{tab:RQ1_results} that the scores for DES are relatively low, and model performance is sentence-dependent, in contrast to the EMO-DB and EmoTale. 
This could be explained by DES being designed differently from the other datasets. 
For example, the sentence ID \texttt{NO} refers to a single word {\it Nej} (No), which may not be sufficient for the model to recognize the emotion. 
Similarly, the sentences with ID: \texttt{SE4}, \texttt{SE5}, \texttt{SE6}, and \texttt{SE8} are all questions, and might be spoken with a different intonation. 
Embeddings from the PTMs generally produce more stable results (low variation), however, a relatively high standard deviation is observed for \texttt{w2v2-FT-dim} features across EmoTale speakers (12.4) and DES sentences (12.7). 
This could be explained by differences inherent in the two datasets: 
EmoTale has a larger age range of speakers compared to the other datasets, and DES contains sentences that vary in linguistic and paralinguistic content.
The UAR scores for Urdu are very speaker-dependent compared to the other datasets.
This can be explained by a high number of speakers and a low number of sentences per speaker.


\begin{figure}
\centering
\includegraphics[width=0.9\columnwidth]{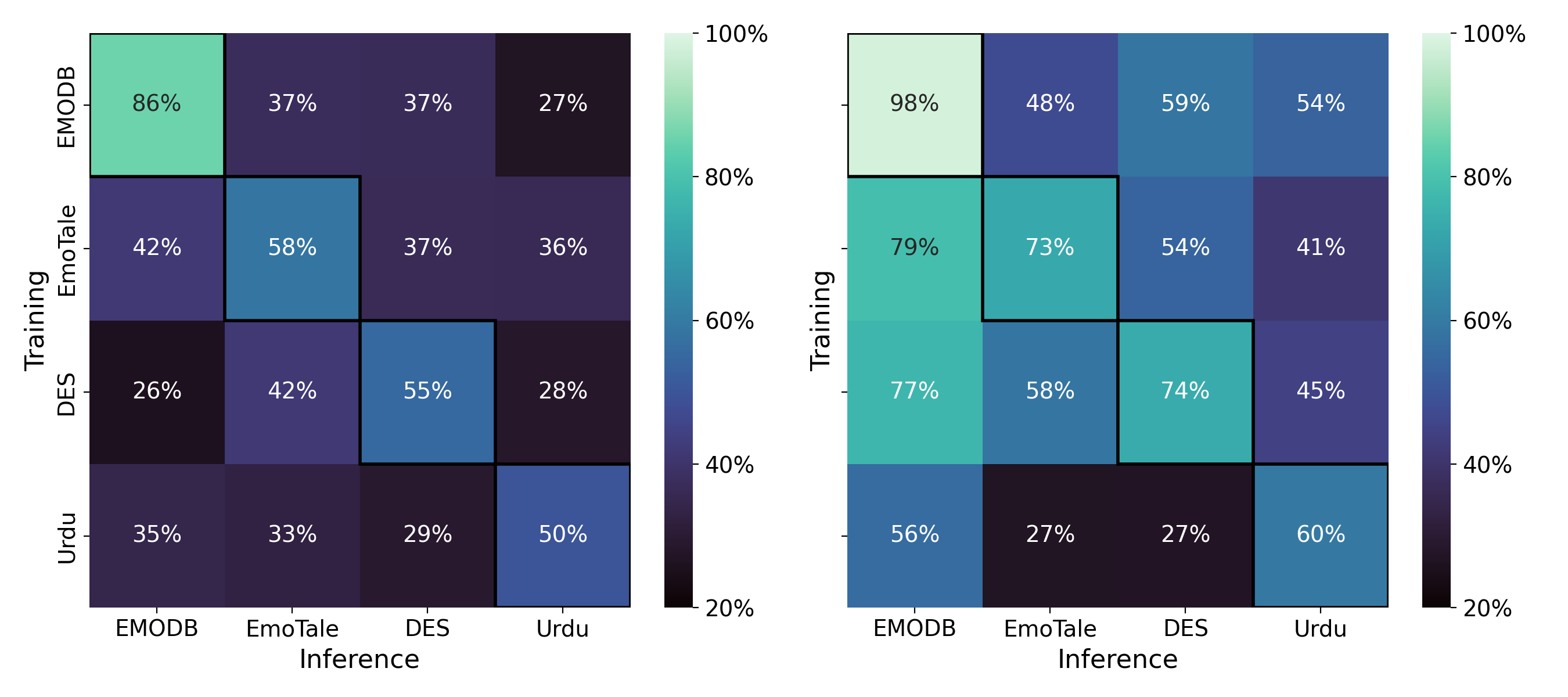}
\caption{Unweighted Average Recall (UAR) scores for SVC trained on ComParE features (left) and pre-trained model embeddings (right). 
}
\label{fig:UAR_trans_confusion}
\vspace{-0.5cm}
\end{figure}

To assess cross-corpus transferability, the SVC models were retrained to recognize a subset of four emotion classes (happy, angry, sad, and neutral) on ComParE and \texttt{w2v2-FT-dim} features for the Emo-DB, EmoTale, DES, and Urdu corpora. 
These datasets were selected specifically because they include all four emotion labels. 
The UAR scores for all train-test combinations are shown in the heatmaps in Fig.~\ref{fig:UAR_trans_confusion}. 
The in-corpus UAR scores in the diagonal of the matrices are found by LOSO cross-validation following the same methodology as earlier, but only including the four emotions.
We wish to develop SER models that generalize well on new, unseen data, especially in real-world applications.
Furthermore, a model that transfers well is less likely to be overfitted on the training data. 
Although performance generally drops in the cross-corpus domain, the deep features seem to be more transferable. 
Importantly, EmoTale proves to be a strong evaluation benchmark. While models trained on EmoTale perform comparably to those trained on other corpora, EmoTale consistently supports meaningful generalization.
For example, inferring on Emo-DB yields higher cross-domain scores than in-corpus UAR scores when trained on EmoTale and DES. 
This continues the pattern from the previous analysis, where Emo-DB achieved significantly higher model performances than the other corpora.
This could be explained by the perception tests carried out during the creation of Emo-DB, where utterances recognized by more than 80\% of the listeners were kept in the database.
Hence, the database is expected to contain utterances with highly pronounced affect.

\section{Conclusions}
Unavailability of Danish affect datasets not only impedes the development of the technology, but also impacts the validation of existing methods on Danish speakers.
We present {\it EmoTale}, a bilingual enacted speech-emotion dataset in Danish and English, intended to enable the evaluation of SER models in the Danish language. In addition to categorical emotion labels, EmoTale includes dimensional annotations for arousal, valence, and dominance. Annotation reliability is high: Concordance Correlation Coefficient (CCC) scores indicate moderate to strong agreement for arousal and valence, and moderate agreement for dominance, while Cohen’s Kappa values indicate substantial consistency in categorical labeling. To demonstrate the \textit{validity} of the dataset, we evaluate its labels and predictive capacity using both pre-trained model embeddings and hand-crafted, acoustic features. Our experiments demonstrate that (a) model performance on EmoTale is comparable to that on established reference datasets, and (b) feature embeddings from PTMs consistently outperform hand-crafted features, particularly in cross-corpus transfer scenarios. While models trained on EmoTale perform comparably to those trained on other corpora, EmoTale consistently supports meaningful generalization. These findings further strengthen the validity of EmoTale as a reliable benchmark for Danish emotional speech.

\section{Acknowledgment}
Co-funded by the French National Research Agency under the Pantagruel project (ANR-23-IAS1-0001) and the European Union under the Marie Skłodowska-Curie Grant Agreement No 101081465 (AUFRANDE). Views and opinions expressed are however those of the author(s) only and do not necessarily reflect those of the European Union or the Research Executive Agency, which cannot be held responsible for them.

\newpage
\section*{Datasheet for Emotale}
\noindent
In line with the proposal on {\it datasheets for datasets} by Gebru et al.~\cite{gebru2021datasheetsdatasets}, we provide the datasheet for the EmoTale corpus, also available as a standalone document with the dataset.

\subsection{Motivation}
\noindent
{\bf For what purpose was the dataset created?}\\
Unavailability of Danish affect datasets not only impedes the development of the technology, but also impacts the validation of existing methods on Danish speakers.
The introduction of our corpus is necessary to, at the very least, be able to validate the performance of SER models for the Danish language.

\noindent
{\bf Who created the dataset and on behalf of which entity?}\\
The dataset was created by Maja Jønck Hjuler, Line Katrine Harder Clemmensen, and Sneha Das at the Technical University of Denmark.\\
\noindent
{\bf Who funded the creation of the dataset?}\\
The dataset creation is funded by the larger WristAngel project which is funded by an exploratory Synergy grant from the Novo Nordisk Foundation and is a collaboration with Copenhagen University Hospital, the Child Psychiatry Research Unit.\\
\subsection{Composition}
\noindent
{\bf What do the instances that comprise the dataset represent?}\\
The instances are audio files of enacted emotional speech in Danish and in English. 
The speakers enact predefined sentences while expressing predefined emotions. 

\noindent
{\bf How many instances are there in total?}\\
The EmoTale corpus consists of a total of 800 audio instances, comprising 450 emotional speech recordings in Danish and 350 in English. Each recording features one of five different enacted emotions, and the dataset is balanced across these emotions.


\noindent
{\bf What data does each instance consist of?}\\
Each instance consists of raw audio data in WAV format, captured at a sampling frequency of 48 kHz. Each recording corresponds to one of five enacted emotions: Neutral, Anger, Sadness, Happiness, or Boredom, and is based on predefined sentences that are translations from the German Emo-DB corpus, designed to be emotionally neutral to minimize contextual bias. 

\noindent
{\bf Is there a label or target associated with each instance?}\\
In addition to the enacted emotion, three independent annotators provided four labels per instance: one categorical for the emotion chosen from the five possible classes, and three numerical for arousal, valence, and dominance in a range of 1 to 5 with increments of 0.5. The ranges are defined as: Valence [1-negative, 5-positive], activation [1-calm, 5-excited], and dominance [1-weak, 5-strong].

\noindent
{\bf Is any information missing from individual instances?}\\
Everything is included. No data is missing.

\noindent
{\bf Are there recommended data splits?}\\
There are no recommended data splits for training, validation, and testing within the dataset itself. However, it is common practice to create stratified splits across speakers and emotions.

\noindent
{\bf Are there any errors, sources of noise, or redundancies in the dataset?}\\
See preprocessing below.


\noindent
{\bf Does the dataset contain data that might be considered confidential?} \\
The data does not contain any signals reflecting on the state of an individual, minimizing the potential negative impact on the individuals. 


\noindent
{\bf Does the dataset identify any subpopulations?}\\
Participants range in age from 9 to 39 years. The dataset includes 18 participants, with 12 females and 6 males.

\noindent
{\bf Is it possible to identify individuals, either directly or indirectly from the dataset?}\\
Individuals can be identified indirectly from the EmoTale corpus due to the unique characteristics of each participant's voice, which can reveal their identity. All participant information has been pseudoanonymized by assigning random IDs. 



\subsection{Collection Process \& Preprocessing}
\noindent
{\bf How was the data associated with each instance acquired?}\\
The data recordings were performed in several sessions in different locations. In each session, the participant was placed in a quiet room and fitted with wireless RØDE microphones paired with the corresponding receiver. Five sentences were enacted with five different emotions, and the participant enacted all sentences for a specific emotion before moving on to the next. The participants were allowed to repeat sentences as often as they liked, but only the last recording was kept in the database. Most often, the recording was made in the first attempt.

\noindent
{\bf Who was involved in the data collection process?}\\ Participants with acting experience and Danish and English language skills were recruited through physical flyers and posts on social media, and theater schools in the Greater Copenhagen area were contacted by email and phone. 

\noindent
{\bf Over what timeframe was the data collected?}\\
The data was collected as part of a master's thesis project lasting 5 months.

\noindent
{\bf Were any ethical review processes conducted?}\\ Ethical approval was obtained from the institutional review board prior to the study~\cite{IRB_emotale}.

\noindent
{\bf Did the individuals in question consent to the collection and use of
their data?}\\ Abiding by GDPR requirements, written consent was obtained from participants or their guardians prior to data collection.


\noindent
{\bf Was any preprocessing/cleaning/labeling of the data done?} \\
\noindent
Some instances were cropped to exclude audible 'clicks' from the experimenter pressing the keyboard at the beginning or end of recordings. The audio files are named according to the same template including information about the language, speaker ID, emotion, and sentence. For example, the file \textit{DK\_004\_A\_5.wav} contains the fifth sentence spoken by speaker 004 in Danish, with angry affect.

\noindent
{\bf Was the “raw” data saved in addition to the preprocessed/cleaned/labeled data?}\\ Yes. The authors can provide the raw data upon request.

\subsection{Uses}
\noindent
{\bf Has the dataset been used for any tasks already?}\\
The dataset paper investigates the dataset's capacity for predicting speech emotions through the development of speech emotion recognition models using Self-Supervised Speech Model embeddings and the openSMILE feature extractor. 
Furthermore, cross-corpus transferability of the models was investigated.


\noindent
{\bf What (other) tasks could the dataset be used for?}\\
The dataset can also be used for ASR, due to the availability of speech and the corresponding transcription. The enacted English speech in addition to Danish will aid research and investigation into speech systems, for instance when the speaker remains identical, but language changes, hence towards more universal speech emotion models.

\noindent
{\bf Are there tasks for which the dataset should not be used?} \\
Given the size of the dataset, it should not be used for tasks that require large-scale training of complex machine learning models. Additionally, it is not suitable for tasks that require spontaneous emotional speech, as the recordings consist of enacted emotions rather than natural emotional expressions.

\subsection{Distribution \& Maintenance}

\noindent
{\bf How will the dataset will be distributed?}\\ The dataset can be accessed at \url{https://github.com/snehadas/EmoTale}.

\noindent
{\bf Will the dataset be distributed under a copyright or other intellectual property (IP) license?} \\ The data will be distributed under a copyright. There is no license, but users are requested to cite the corresponding paper if the dataset is used.

\noindent
{\bf Who will be maintaining the dataset and how can they be contacted?} \\
The dataset will be maintained by the corresponding author Sneha Das~(sned@dtu.dk).

\noindent
{\bf Will the dataset be updated?} \\ This dataset will not be updated in terms of the number of samples or participants.


\clearpage
\bibliographystyle{IEEEtran}

\bibliography{camera_ready}

\end{document}